\documentclass[11pt]{article}

\usepackage[final]{acl}

\usepackage{times}
\usepackage{latexsym}

\usepackage{tcolorbox} %22222222
\usepackage{tcolorbox}
\tcbuselibrary{breakable}
\usepackage{makecell} 
\usepackage{amsmath}

\usepackage[T1]{fontenc}
\usepackage[utf8]{inputenc}

\usepackage{microtype}

\usepackage{inconsolata}

\usepackage{graphicx}

\title{CLaC at SemEval-2026 Task 6: Response Clarity Detection\\in Political Discourse}
\author{Nawar Turk, Lucas Miquet-Westphal, Leila Kosseim \\
  Computational Linguistics at Concordia (CLaC) Lab \\
  Dept. of Computer Science and Software Engineering \\
  Concordia University, Montréal, Québec, Canada \\
  \texttt{\{nawar.turk, lucas.miquet-westphal\}@mail.concordia.ca}, \\
  \texttt{leila.kosseim@concordia.ca}}

\begin{document}
\maketitle
\begin{abstract}
In this paper, we present our system for SemEval-2026 Task~6 (CLARITY) on response clarity and evasion detection in question-answer pairs from U.S. presidential interviews, comparing fine-tuned encoders with prompt-based LLMs. Our LLM ensemble achieves $80$ macro-F1 on the $3$-class Task~1 ($9^{\text{th}}\!/41$) and $59$ on the $9$-class Task~2 ($3^{\text{rd}}\!/33$). Across $8$ transformer encoders optimized through a four-stage pipeline, partial encoder layer unfreezing outperforms full fine-tuning by a wide margin. Combining English and multilingual encoders further improves ensemble performance over either family alone, despite multilingual models being individually weaker. Prompt-based LLMs, without any task-specific parameter updates, outperform fine-tuned encoders, particularly on minority classes; among open-weight LLMs, parameter count does not predict performance. Enriched input, concatenating the full interviewer turn, improves LLM performance but not that of encoders, an effect that persists with Longformer's extended context window, suggesting the divergence is not attributable to sequence-length capacity alone in our settings. The \textit{Clear Reply}/\textit{Ambivalent} boundary remains the dominant failure mode, mirroring the disagreement among human annotators. Our code, prompts, model configurations, and results are publicly available.\footnote{\url{https://github.com/CLaC-Lab/SemEval-2026-task6-CLARITY}}

\end{abstract}

\section{Introduction}
Politicians typically avoid direct answers in interviews \citep{bull2003}, hence tools for detecting evasion are valuable for political discourse research and accountability. The SemEval-2026 CLARITY task \citep{thomas2026semeval2026task6clarity} addresses the automatic detection of response clarity and evasion in English question-answer pairs from U.S. presidential interviews, building on the dataset and taxonomy introduced by \citet{thomas-etal-2024-never}. Task~1 requires classifying responses into one of 3 Clarity categories, while Task~2 predicts one of 9 fine-grained Evasion labels. We participated in both tasks (with a focus on Task~1) exploring 3 approaches: (1)~encoder-based models optimized through a four-stage pipeline, (2)~a Longformer-based architecture to account for long-context instances, and (3)~prompt-based LLMs evaluated across multiple prompting strategies with top configurations combined into a final ensemble. Due to a lack of time, the best-performing Task~1 configuration was applied to Task~2. Overall, our LLM ensemble achieves $80$ macro-F1 on the Task~1 test set and $59$ on the Task~2 test set, outperforming fine-tuned encoders.

\section{Background} 
\subsection{Dataset Description}
\begin{figure}[h]
\centering
\includegraphics[width=\columnwidth]{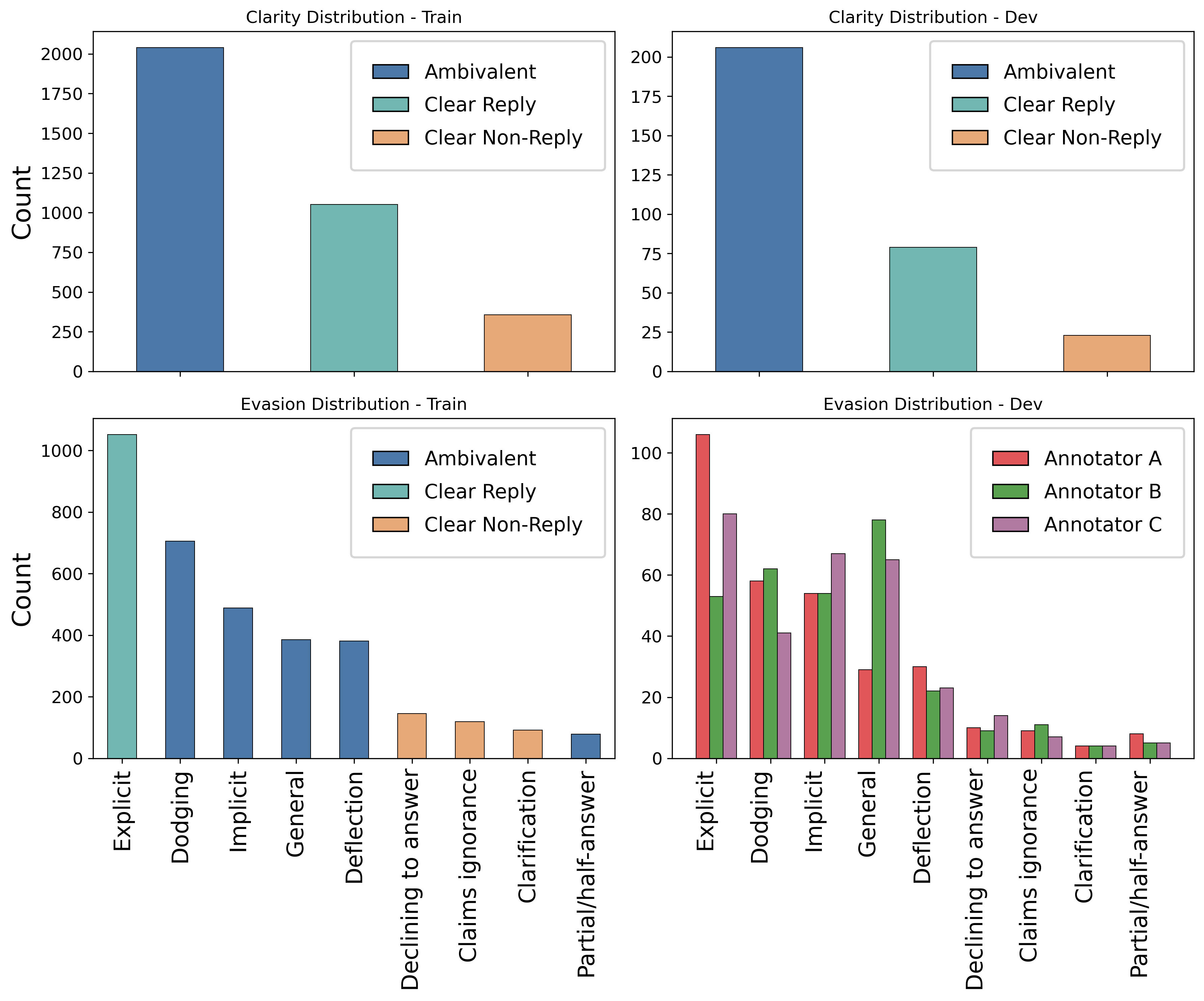}
\caption{Label distributions across Clarity (top) and Evasion (bottom) levels for the train and dev sets.}
\label{fig:label_distributions}
\end{figure}

The dataset provided by the organizers consists of 3,993 English question-answer pairs extracted from televised U.S. presidential interviews, split into training (3,448 instances), dev (308), and test (237) sets. Each instance is annotated at two levels: 3 Clarity labels (Level~1) and Evasion labels (Level~2) that subcategorize the Clarity labels into 9 more fine-grained categories. Figure~\ref{fig:label_distributions} shows label distributions across both levels for the train and dev sets; train Evasion labels are color-coded by their parent Clarity class, while dev Evasion labels are shown per annotator. 
The training set provides a single gold label per level; while the dev set includes one gold Clarity label but 3 independent Evasion annotations per instance. The visible annotator disagreement highlights the inherent difficulty of fine-grained Evasion classification. Both levels exhibit substantial class imbalance: \textit{Ambivalent} dominates Level~1, while Level~2 is heavily skewed toward \textit{Explicit} and \textit{Dodging}.

\subsection{Related Work}
Political evasion has been widely studied in political science \citep{RASIAH2010664, bull2003}, with \citet{bull2003} finding that politicians frequently avoid directly answering interview questions. \citet{thomas2026semeval2026task6clarity, thomas-etal-2024-never} formalize this phenomenon computationally through the CLARITY task\footnote{\url{https://konstantinosftw.github.io/CLARITY-SemEval-2026/}}, a two-level hierarchical taxonomy, and a dataset of political question-answer pairs annotated through a combination of human expertise and LLM-assisted validation, alongside baseline evaluations across multiple architectures. Building on this foundation, we investigate how encoder adaptation techniques and prompting strategies influence performance across Clarity classes, particularly under class imbalance and the hierarchical structure of the labels.

\section{System Overview}

We address Task~1 using 3 approaches: (1)~encoder-based models, (2)~Longformer-based architecture, and (3)~prompt-based LLM classifiers. Among all Task~1 experiments, the LLM prompting approach with 27 shots achieved the strongest performance. Due to a lack of time, we therefore adopt this strategy for Task~2.
\begin{figure}[t]
\centering
\includegraphics[width=\columnwidth]{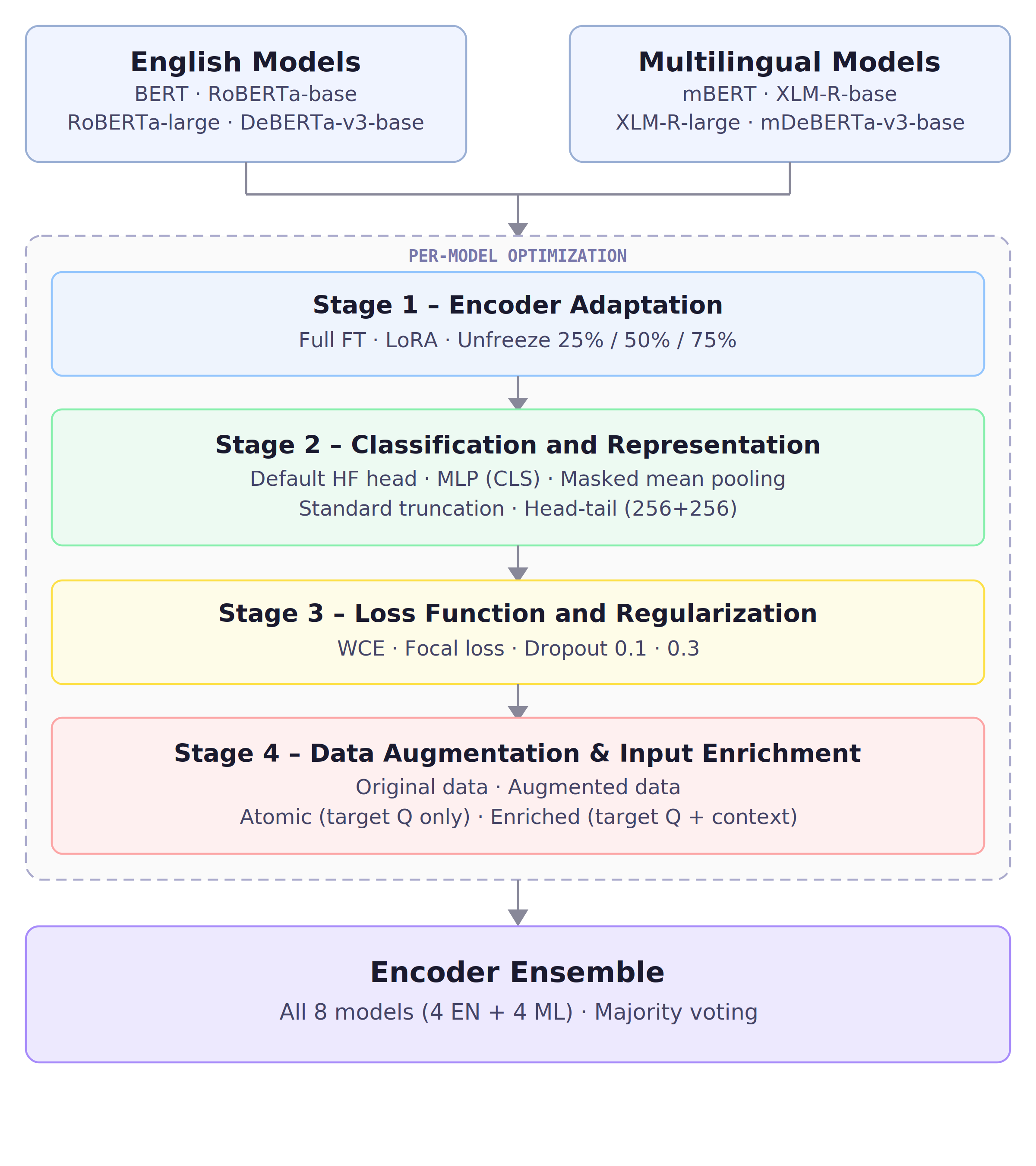}
\caption{Encoder pipeline. Each model is optimized independently across stages before final ensembling.}
\label{fig:System1-encoders}
\end{figure}

\subsection{Encoder-Based Models}
\subsubsection{Models}

We evaluate 8 transformer encoders spanning English and multilingual variants in base and large configurations. English models include \texttt{bert-base-\allowbreak uncased}
~\citep{devlin-etal-2019-bert},
\texttt{RoBERTa-\allowbreak base}, \texttt{RoBERTa-\allowbreak large}~\citep{liu2019robertarobustlyoptimizedbert},
and \texttt{DeBERTa-\allowbreak v3-\allowbreak base}
~\citep{he2023debertav}. Multilingual counterparts include
\texttt{bert-\allowbreak base-\allowbreak multilingual-\allowbreak cased},
\texttt{xlm-\allowbreak RoBERTa-\allowbreak base}, \texttt{xlm-\allowbreak RoBERTa-\allowbreak large}
~\citep{conneau-etal-2020-unsupervised},
and \texttt{mDeBERTa-\allowbreak v3-\allowbreak base}. We include multilingual models to assess whether cross-lingual pretraining improves model performance.

% [[ ADD ME IF YOU HAVE SOME SPACE, THank YoU]]
% This setup allows us to compare different architectures, pre-training data scope (English vs. multilingual), and model size. 
    
\subsubsection{Training Strategy}\label{sec:system}
Figure~\ref{fig:System1-encoders} illustrates the training strategy of the encoder models where each model is optimized through 4 sequential stages. At each stage, we evaluate multiple configurations and carry forward the variant achieving the highest macro-F1.  The input to all encoder models is the concatenated QA pair. To mitigate label imbalance, we apply class-weighted cross-entropy (WCE) (see Appendix~\ref{app:loss_weighting}). We use a dropout rate ($0.1$) for regularization. In later stages, we experiment with alternative loss functions, dropout values and regularization strategies.

% To address label imbalance, we apply class-weighted cross-entropy in the initial stages, where $N$ is the total number of training instances, $C$ the number of classes, and $n_i$ the number of instances in class $i$.
% \[w_i = \frac{N}{C \cdot n_i}\]

\paragraph{Stage 1: Encoder Adaptation.} We compare (a) full fine-tuning, (b) LoRA-based tuning with a frozen encoder backbone, and (c) partial unfreezing of the top 25\%, 50\%, or 75\% layers.

\paragraph{Stage 2: Classification \& Representation.} We evaluate 3 classification heads (default: a single linear layer over CLS; MLP over CLS; and mean pooling with a single linear layer; see Appendix~\ref{app:heads}), each under standard truncation (preserving the first $512$ tokens) and head-tail truncation (preserving the first and last $256$ tokens). 

\paragraph{Stage 3: Loss Function \& Regularization.} We additionally 
evaluate focal loss and dropout $0.3$ against the WCE + dropout $0.1$ baseline.
% Building on the best encoder adaptation and representation configuration selected over Stages~1 and~2 for each model, we explore alternative loss functions and dropout rates. Using WCE as the default loss and a dropout rate of 0.1 as the baseline regularization, we additionally evaluate focal loss and a higher dropout rate of 0.3.

\paragraph{Stage 4: Data Augmentation \& Input Enrichment.} 
We evaluate two dimensions: (a)~training data, comparing the original dataset with an augmented version that upsamples \textit{Clear Non-Reply} via LLM-generated paraphrases (see Appendix~\ref{app:augmentation}); and (b)~input mode, comparing atomic input (target question only) against enriched input incorporating the full interviewer turn 
(see Appendix~\ref{app:enriched}).\footnote{The full interviewer turn 
may contain multiple questions; we refer to this as \textit{context} 
throughout.}

\paragraph{Ensemble Strategy.}
After Stage~4, we retain the configuration achieving the highest dev macro-F1 for each of the 8 encoder models and construct an ensemble using majority voting. In case of a tie, we select the label with the highest mean predicted probability score among the tied candidates.

\subsection{Long-Context Model}
Since approximately 30\% of the training QA pairs exceed the 512-token limit, we experiment with \texttt{longformer-base-4096} to assess whether long-context modelling improves performance.

\paragraph{Stage 1: Configuration Selection.} We base our initial design on \texttt{DeBERTa-v3-base}, the best-performing model across all encoder stages, and we adopt its best configuration. We then conduct early exploration with alternative configurations (different classification heads, dropout $0.3$, varying max lengths). Our exploration did not improve macro-F1 on the dev set; we therefore retained the original configuration.

\paragraph{Stage 2: Ablation Study.} With the core configuration fixed, we systematically evaluate 3 design dimensions across 12 configurations: classification head type (default vs.\ MLP), input ordering (3 scenarios: context-question-answer, question-context-answer, question-context-answer-question-repeat), and global attention pattern (CLS-only or extended to question tokens). The maximum sequence length was set to 2048, with a 512-token attention window (see Appendix~\ref{app:longformer-configs}).

\subsection{LLM-Based Models}
\subsubsection{Models}
We evaluate both open-weight and proprietary large language models. 
Open-weight families include LLaMA (\texttt{LLaMA-\allowbreak 3.1-\allowbreak Nemotron-\allowbreak Ultra-\allowbreak 253B} \citep{bercovich2025llamanemotronefficientreasoningmodels}; \texttt{LLaMA-\allowbreak 3.3-\allowbreak 70B-\allowbreak Instruct} \citep{grattafiori2024llama3herdmodels}), Qwen (\texttt{Qwen3-\allowbreak 235B-\allowbreak Instruct}; \texttt{Qwen3\allowbreak-\allowbreak 80B-\allowbreak Instruct}; \texttt{Qwen3-\allowbreak 30B-\allowbreak Instruct}) \citep{yang2025qwen3technicalreport}, and Mixtral (\texttt{Mixtral-\allowbreak 8x22B-\allowbreak Instruct}). 
Among proprietary models, we evaluate \texttt{GPT-5}~\citep{singh2025openaigpt5card} (OpenAI) across the full prompt sweep. After identifying the best prompting strategy, we evaluate \texttt{Claude-Opus-4.5} (Anthropic) and \texttt{Gemini-3-Flash-Preview} (Google) under this strategy for prompt optimization and final comparison. The corresponding checkpoints for all open-weight models are provided in Appendix~\ref{app:llm-checkpoints}.
\subsubsection{Task~1 Prompting Strategy}
\label{sec:task1-prompting}

\paragraph{Stage 1: Technique \& Model Sweep.}
Our prompting design is informed by the survey of prompt engineering techniques of~\citet{schulhoff2025promptreportsystematicsurvey}. 
We evaluate zero-shot (ZS), zero-shot with instruction repeated (ZS+Re2)~\citep{xu-etal-2024-reading}, few-shot prompting with 3, 9 and 27 shots (FS3/FS9/FS27), and a prompt inspired by chain-of-thought (CoT) prompting~\citep{NEURIPS2022_8bb0d291}; however, we instruct the model to reason internally before producing the final label.
Few-shot demonstrations are class-balanced across the 3 Level-1 labels (multiples of 3). All strategies are evaluated under both atomic and enriched input settings. Prompt templates are provided in Appendix~\ref{app:prompts}.

\paragraph{Stage 2: Prompt Optimization \& Model Expansion.} Building on FS27 with enriched input (target question concatenated with the full interviewer turn), the best-performing prompt during Stage 1, we conduct prompt optimization by introducing two controlled modifications: (a) prepending the president's name, and (b) augmenting each Clarity label definition with its Level-2 Evasion subcategories, while still predicting only the Level-1 label. The refined prompt is then extended to \texttt{Claude-\allowbreak Opus-\allowbreak 4.5} and \texttt{Gemini-\allowbreak 3-\allowbreak Flash-\allowbreak Preview} to assess cross-model robustness.
\paragraph{Ensemble Strategy.} After prompt optimization and model expansion, we select the top 3 configurations based on dev macro-F1: \texttt{GPT-\allowbreak 5}, \texttt{Gemini-3-Flash-Preview}, and \texttt{Qwen3-\allowbreak 235B}, all under the FS27 enriched setting with subcategory-augmented definitions. Gemini and Qwen additionally incorporate president-name conditioning, while \texttt{GPT-\allowbreak 5} does not as it performed better without it. Predictions are combined via majority voting, defaulting to \textit{Ambivalent} on ties. 

% \paragraph{Binary Prompting Variant.} We additionally explored a one-vs-rest binary prompting formulation, 
% where three independent Yes/No classifiers (one per Level-1 label) are aggregated via majority voting. This variant did not outperform 
% the direct multi-class formulation; full details are provided in Appendix~\ref{app:binary}.

\subsubsection{Task~2 Prompting Strategy}
For Task~2, we did not perform a separate prompt sweep due to a lack of time. Instead, we used the best-performing ensemble configuration from Task~1: \texttt{GPT-5}, \texttt{Gemini-\allowbreak 3-\allowbreak Flash-\allowbreak Preview}, and \texttt{Qwen3-\allowbreak 235B-\allowbreak Instruct}, all under FS27 using enriched input and subcategory-augmented definitions. Gemini and Qwen additionally incorporate president-name conditioning, while GPT does not. Predictions are combined via majority voting. In case of ties, we default to the class with the highest frequency among the tied classes in the training set.

% [[ ADD ME IF THERE IS SPACE ]]
% We explored two inference variants: (1) direct prediction over all nine fine-grained subclasses; and (2) constrained prediction, where the model selects only from the subclasses corresponding to the Task~1 majority-voted class.

\section{Experimental Setup}
Encoder and Longformer models are implemented in PyTorch using Hugging Face Transformers. All models are trained on the official training split and selected based on dev macro-F1. Class imbalance is addressed via weighted cross-entropy (Appendix~\ref{app:loss_weighting}). LLM-based experiments are conducted via official APIs, with prompt templates and few-shot demonstrations held fixed across models. Complete hyperparameters, hardware setup, Longformer configurations, and prompt templates are provided in Appendices~\ref{app:hyperparams}, \ref{app:longformer-configs}, and~\ref{app:prompts}.

\section{Results \& Analysis} Table~\ref{tab:main_results} reports Task~1 dev macro-F1 for each encoder using its best configuration after the four-stage optimization process (see Section~\ref{sec:system}). For LLM-based systems, we report the top-performing models retained for ensembling after the two-stage prompting process (see Appendix~\ref{app:llm-stage1} for the best configuration of the remaining LLMs). Among encoder-based models, \texttt{DeBERTa-v3-base} performs best ($65.1$), and ensembling the 8 optimized encoders improves performance to $70.5$ on the dev set; while the best-performing Longformer configuration achieves $64.3$. LLM-based systems outperform fine-tuned encoders, with \texttt{Gemini-\allowbreak 3-\allowbreak Flash-\allowbreak Preview} reaching $71.9$. The LLM ensemble, used for both our Task~1 and Task~2 submissions, achieves $78.1$ on the dev set and $80$ on the test set. Applying this configuration to Task~2 yields $59$ macro-F1 on the test set.

\begin{table}[t]
\centering
\small
\setlength{\tabcolsep}{2.5pt}
\begin{tabular}{llc}
\hline
\textbf{Design Choice} & \textbf{Avg F1 ($\Delta$)} & \textbf{\# Sel.} \\
\hline

\multicolumn{3}{l}{\textbf{Stage 1: Encoder Adaptation}} \\
\textit{Full Finetuning} & \textit{41.2} & 0 \\
Unfreeze Top 25\% & \textbf{59.1 (+17.9)} & \textbf{6} \\
LoRA (r = 16) & 56.1 (+14.9) & 1 \\
Unfreeze Top 75\% & 45.1 (+3.9) & 1 \\
\hline

\multicolumn{3}{l}{\textbf{Stage 2: Classification \& Representation}} \\
\textit{Def. Head + Std. Trunc.} & \textit{60.1} & 2 \\
Def. Head + Head-Tail Trunc. & 59.9 (-0.2) & 2 \\
MeanPool + Head-Tail Trunc. & 59.9 (-0.2) & 1 \\
MeanPool + Std. Trunc. & 59.8 (-0.3) & 2 \\

MLP Head + Head-Tail Trunc. & 57.8 (-2.3) & 1 \\
\hline

\multicolumn{3}{l}{\textbf{Stage 3: Loss \& Regularization}} \\
\textit{WCE + Dropout = 0.1 }& \textit{61.5} & 8 \\
\hline

\multicolumn{3}{l}{\textbf{Stage 4: Data Augmentation \& Input Enrichment}} \\
\textit{Original Data + Atomic Input} & \textit{61.5} & 5 \\
Augmented Data + Enriched Input & 59.8 (-1.7) & 2 \\
Augmented Data + Atomic Input & 59.8 (-1.7) & 1 \\
Original Data + Enriched Input & 59.1 (-2.4) & 0 \\
\hline
\end{tabular}
\caption{Average dev macro-F1 across 8 encoder models per design choice. $\Delta$ is relative to the stage baseline, and \# Sel. is the number of best models selecting that choice. \textit{Italicized rows} indicate the stage baseline.}
\label{tab:ablation_stages}
\end{table}

\begin{table}[t]
\centering
\small
\setlength{\tabcolsep}{4pt}
\begin{tabular}{llc}
\hline
\textbf{System Type} & \textbf{Model} & \textbf{Macro-F1} \\
\hline
Encoder-Based & \textbf{\texttt{DeBERTa-v3-base}} & \textbf{65.1} \\
            & \texttt{RoBERTa-base} & 63.8 \\
            & \texttt{BERT-base} & 63.4 \\
            & \texttt{RoBERTa-large} & 63.0 \\
            & \texttt{mBERT-base} & 62.9 \\
            & \texttt{xlm-RoBERTa-base} & 62.6 \\
            & \texttt{xlm-RoBERTa-large} & 61.4 \\
            & \texttt{mDeBERTa-\allowbreak v3-\allowbreak base} & 60.8 \\
\cline{2-3}
          & Ensemble (EN only) & 65.6 \\
            & Ensemble (Multi only) & 68.2 \\
            & \textbf{Ensemble (EN+Multi)} & \textbf{70.5} \\\hline
Long Context  & \texttt{Longformer-base-4096} & 64.3 \\
\hline
LLM-Based     & \textbf{\texttt{Gemini-3-Flash-Preview}} & \textbf{71.9} \\
              & \texttt{GPT-5} & 71.5 \\
              & \texttt{Qwen3-235B-Instruct} & 68.3 \\
\cline{2-3}
              & \textbf{LLM Ensemble} & \textbf{78.1} \\
\hline
\end{tabular}
\caption{Task~1 dev macro-F1 across system approaches. Each row reports the best config after full optimization.}
\label{tab:main_results}
\end{table}

\subsection{Encoder and Long-Context Results}
Table~\ref{tab:ablation_stages} reports, for each encoder design choice, the average dev macro-F1 across the 8 encoder models, its change ($\Delta$) relative to the stage baseline, and the number of models that selected that choice in their optimal configuration. For each stage, the baseline is the average dev macro-F1 of the best configuration carried over from all previous stages, except for Stage~1, where full fine-tuning serves as the baseline.

Table~\ref{tab:ablation_stages} shows that encoder adaptation is a critical optimization step: partially unfreezing the top 25\% layers improves macro-F1 by $+17.9$ over full fine-tuning ($59.1$ vs. $41.2$), suggesting that full fine-tuning may overfit the limited training size. Subsequent stages build on the best configuration of the previous stage but yield only marginal changes. In Stage~2, no classification head or truncation strategy consistently improves across all models over the carried-forward baseline; in Stage~3, all 8 models selected weighted cross-entropy with dropout $0.1$ as their optimal configuration, indicating that class weighting is sufficient without additional loss reshaping. In Stage~4, data augmentation decreases average performance across models ($61.5$ to $59.8$ macro-F1). However, its effect is model-dependent: 3 models, \texttt{DeBERTa-v3-base}, \texttt{mBERT-base}, and \texttt{xlm-\allowbreak RoBERTa-\allowbreak base}, improve under at least one augmented configuration (see Appendix~\ref{app:encoder-configs}), suggesting that paraphrase-based augmentation does not generalize uniformly across architectures.

As shown in Table~\ref{tab:main_results}, ensembling the 8 optimized encoders yields an additional $+5.4$ macro-F1 ($65.1$ to $70.5$), indicating that architectural diversity provides complementary signal. While English-only encoders achieve higher standalone macro-F1 than multilingual variants, the multilingual-only ensemble ($68.2$) surpasses the English-only ensemble ($65.6$), and combining both families produces the strongest result ($70.5$). This suggests that multilingual pretraining produces models with distinct decision boundaries, which improve overall performance when combined with English-only encoders.
The Longformer ($64.3$) fell below the encoder ensemble despite its extended context capacity (see Appendix~\ref{app:longformer-configs} for further details).

\subsection{LLM-Based Results}
\begin{figure}[t]
    \centering
    \includegraphics[width=\linewidth]{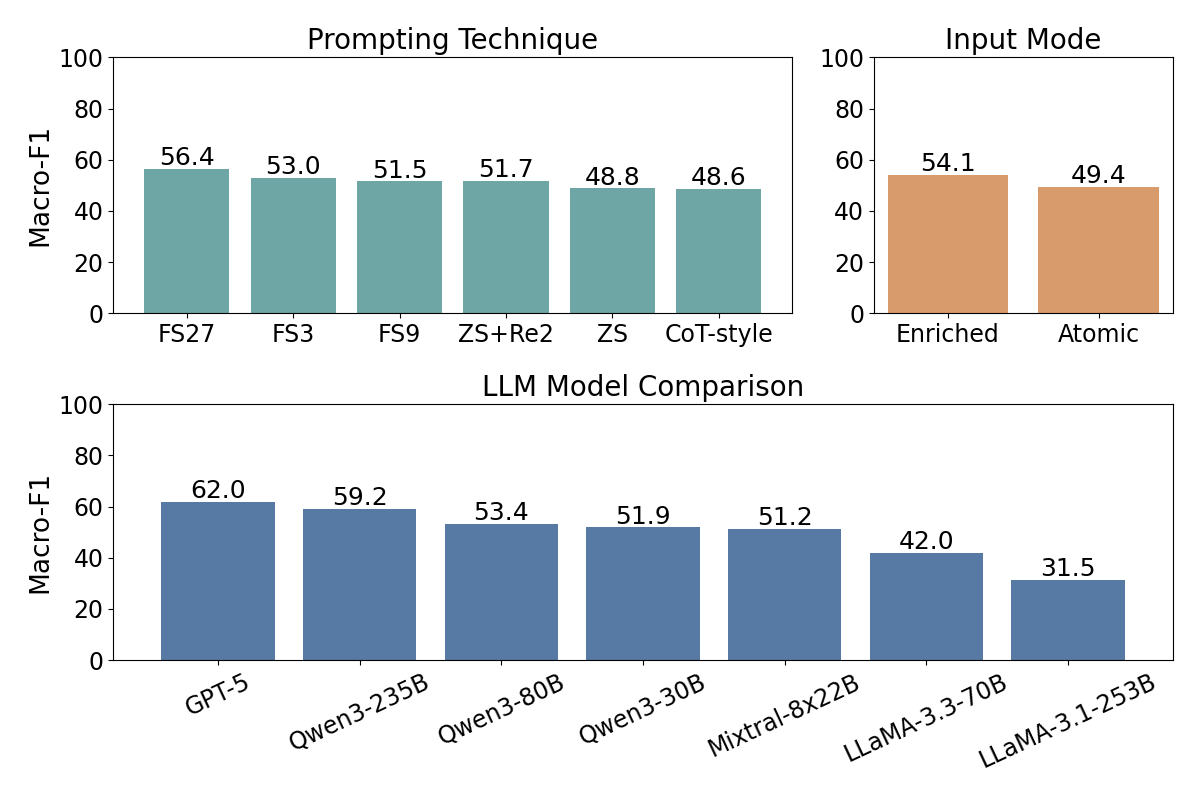}
    \caption{
    Impact of prompting strategy, input configuration, and LLM choice on dev macro-F1. Each panel reports averages over the remaining variables, computed across all experimental configurations. }
    \label{fig:design_vs_model_dev}
\end{figure}

Figure~\ref{fig:design_vs_model_dev} summarizes dev macro-F1 results from the Stage~1 prompting sweep, prior to prompt refinement and model expansion. Each panel reports averages over the remaining variables, computed across all experimental configurations. The top row isolates design choices, comparing prompting strategy (left) and input configuration (right). Few-shot prompting (FS27) achieves the strongest average performance ($56.4$), while the enriched input consistently outperforms atomic input ($+4.7$), indicating consistent gains across configurations from inference-time design choices. The bottom panel shifts focus to LLM choice, revealing larger variation across models: \texttt{GPT-5} reaches $62.0$ macro-F1, whereas \texttt{LLaMA-\allowbreak 3.1-\allowbreak Nemotron-\allowbreak Ultra-\allowbreak 253B} trails at $31.5$. Notably, parameter count alone does not explain these differences. While having additional context leads to measurable improvements, the choice of LLM accounts for substantially larger performance variation in Clarity detection. The best configuration for each model is reported in Appendix~\ref{app:llm-stage1}. 

Prompt refinement in Stage~2 further improves performance across models. For \texttt{GPT-5}, macro-F1 rises from $64.4$ under the best Stage~1 configuration to $71.5$ after refinement ($+7.1$), driven by augmenting the 3 Clarity label definitions with their corresponding Level~2 Evasion subcategory descriptions. This result highlights the impact of prompt engineering and hierarchical label information on performance, without any model fine-tuning.

\subsubsection{Cross-System Comparison}

\begin{table}[t]
\centering
\small
\setlength{\tabcolsep}{2pt}
\begin{tabular}{lccc}
\hline
\textbf{Ensemble} & \textbf{Clear Reply} & \textbf{Ambivalent} & \textbf{Clear Non-Reply} \\
\hline
LLM & \textbf{70.7} & \textbf{85.3} & \textbf{78.4} \\
Encoder & 61.0 & 82.4 & 68.1 \\
\hline
\end{tabular}
\caption{Dev F1 per class for the encoder and LLM ensembles.}
\label{tab:llm_vs_encoder}
\end{table}
Table~\ref{tab:llm_vs_encoder} shows dev F1 per class for the encoder and LLM ensembles. The LLM ensemble improves performance across all classes, with the largest gains on the minority classes, \textit{Clear Reply} ($+9.7$) and \textit{Clear Non-Reply} ($+10.3$). This pattern suggests greater robustness to class imbalance in underrepresented categories. The enriched input reveals system divergence; for the encoder models, adding the full interviewer turn reduced the average macro-F1 (e.g., Original + Enriched: $59.1$, $\Delta = -2.4$; see Table~\ref{tab:ablation_stages}). The Longformer, despite long-context modeling, underperforms standard encoders in our settings. In contrast, LLMs consistently gain from additional context ($54.1$ vs. $49.4$, $\Delta = +4.7$; Figure~\ref{fig:design_vs_model_dev}), suggesting that LLMs leverage enriched context more effectively than encoder architectures, beyond sequence-length capacity alone. Error analysis (Appendix~\ref{app:confusion-matrices}) reveals that both ensembles share the same dominant error pattern: bidirectional confusion between \textit{Clear Reply} and \textit{Ambivalent}, mirroring the lowest pairwise annotator agreement among Clarity classes ($\kappa = 0.65$; \citealt{thomas-etal-2024-never}). The LLM ensemble improves \textit{Clear Non-Reply} recall ($20/23$ vs.\ $16/23$), consistent with its stronger minority class performance seen in Table~\ref{tab:llm_vs_encoder}.

% \section{Conclusion}
% We systematically compared encoder adaptation and prompt-based LLMs for SemEval-2026 CLARITY. Partial unfreezing improves encoder performance, yet inference-time prompting proves more effective than fine-tuning, particularly on minority classes. Ensemble diversity matters more than model scale, with multilingual models contributing better when combined with English-only encoders. The persistent \textit{Clear Reply}/\textit{Ambivalent} confusion mirrors annotator disagreement, reflecting the inherent ambiguity of political discourse.

\section{Conclusion}
We systematically compared encoder adaptation and prompt-based LLMs for SemEval-2026 CLARITY. Our LLM ensemble achieves $80$ macro-F1 on Task~1 ($9^{th}\!/41$) and $59$ on Task~2 ($3^{rd}\!/33$), with prompt refinement and hierarchical label information driving the largest gain. Partial unfreezing improves encoder performance, yet inference-time prompting proves more effective than fine-tuning, particularly on minority classes. Ensemble diversity matters more than model scale, with multilingual models contributing better when combined with English-only encoders. The persistent \textit{Clear Reply}/\textit{Ambivalent} confusion mirrors annotator disagreement, reflecting the inherent ambiguity of political discourse. Together, these findings suggest that for low-resource, hierarchically-labeled tasks like CLARITY, inference-time design choices (prompt structure, label hierarchy, ensemble composition) are more impactful than parameter-level adaptation.

% \section*{Limitations}
% The dataset is restricted to U.S. presidential interviews and contains fewer than 4K instances, which may limit generalization to other political settings or languages. The boundary between \textit{Clear Reply} and \textit{Ambivalent} is inherently subjective, which is reflected in moderate annotator agreement ($\kappa$=0.65, \citealt{thomas-etal-2024-never}). LLM-based results rely partially on proprietary APIs (GPT-5, Gemini), which may affect long-term reproducibility due to model versioning. Our CoT-inspired prompting prevents inspection of intermediate reasoning, limiting interpretability and may have affected results. Finally, our encoder study spans 136 configurations across 8 models and 4 optimization stages; due to computational constraints, each configuration was evaluated once with a fixed random seed, and variance across seeds was not assessed.

\section*{Limitations and Future Work}
The dataset is restricted to U.S. presidential interviews and contains fewer than 4K instances, which may limit generalization to other political settings or languages. The boundary between \textit{Clear Reply} and \textit{Ambivalent} is inherently subjective, which is reflected in moderate annotator agreement ($\kappa = 0.65$; \citealt{thomas-etal-2024-never}). LLM-based results rely partially on proprietary APIs (GPT, Gemini), which may affect long-term reproducibility due to model versioning. Our CoT-inspired prompting prevents inspection of intermediate reasoning, limiting interpretability and may have affected results. Our encoder study spans 136 configurations across 8 models and 4 optimization stages; due to computational constraints, each configuration was evaluated once with a fixed random seed, and variance across seeds was not assessed. Future work should quantify this variance through multi-seed runs on the final per-model configurations to provide stability estimates for the reported encoder results.

Several findings in this paper are reported empirically but not yet causally explained: \textit{why} multilingual encoders underperform individually but strengthen the ensemble, \textit{why} enriched input helps LLMs but degrades encoder performance, and \textit{why} Longformer fails to leverage its long-context capacity in this setting. We frame the latter as observed under our current configurations rather than a general property of long-context models.

Our Task~2 system reuses the optimal Task~1 ensemble configuration without an independent prompt sweep over the 9-class label space. While this transfer resulted in competitive performance ($3^{rd}/33$), it does not establish that the configuration is optimal for the Evasion classification task. Future work should conduct a dedicated Task~2 prompt sweep to determine whether the Task~1 optimum transfers to the 9-class Evasion classification, or whether fine-grained Evasion classification benefits from different prompting strategies.

Finally, we did not perform a head-to-head comparison with the top-ranked Task~6 systems (TeleAI at $89$ macro-F1 on Task~1). Understanding whether the 9-point gap stems from prompt design, model choice, or fundamentally different architectures would contextualize our results; we leave such cross-system analysis to future work once participating systems are publicly described.

\section*{Acknowledgments}
We thank the SemEval-2026 Task~6 organizers for designing and running the CLARITY shared task, and the anonymous reviewers for their constructive feedback. We also thank the Pierre Arbour Foundation, the Natural Sciences and Engineering Research Council of Canada (NSERC), and the Fonds de recherche du Québec (FRQ) for their financial support.
\\
\\
% Bibliography entries for the entire Anthology, followed by custom entries
%\bibliography{anthology,custom}
% Custom bibliography entries only
\bibliography{custom}

\appendix

\section{Optimization and Loss}
\label{app:optimization}

\subsection{Loss Functions}
\label{app:loss_weighting}

To address label imbalance, we assign each class $i$ a weight $w_i$ computed as:
\[
w_i = \frac{N}{C \cdot n_i}
\]
where $N$ is the total number of training instances; $C$, the number of classes; and $n_i$, the number of instances in class $i$. This places greater weight on rarer classes during training. All weights are computed from the training split only. Both loss functions evaluated in Stage~3 incorporate these class weights.

\paragraph{Weighted Cross-Entropy (WCE).} The standard formulation used as the default loss across all stages, applying class weights to the cross-entropy term:

\[
\mathcal{L}_{\text{WCE}} = -w_t \log(p_t)
\] where $p_t$ denotes the predicted probability of the ground-truth class $t$.
\paragraph{Focal Loss.}
Evaluated as an alternative in Stage~3, this formulation adds a focal factor $(1 - p_t)^{\gamma}$ that down-weights well-classified examples to focus training on harder instances:

\[
\mathcal{L}_{\text{Focal}} = -w_t (1 - p_t)^{\gamma} \log(p_t)
\]
where $\gamma = 2.0$.

\subsection{Training and Decoding Configuration}
\label{app:hyperparams}
Table~\ref{tab:hyperparams} reports the training configuration for encoder and Longformer models. For partial unfreezing, the embedding layer and all but the top $k$\% of encoder layers are frozen; the classification head remains trainable throughout. For LLM-based experiments, Hugging Face-hosted models use temperature 0; all other models (GPT, Gemini, Claude) use default API generation settings. All encoder and Longformer experiments were conducted on an NVIDIA RTX 4070 12GB with 1 GPU. Training time per model ranged from approximately 5 minutes to 6 hours. Software dependencies are listed in Table~\ref{tab:deps}.

\begin{table}[h]
\centering
\small
\begin{tabular}{lll}
\hline
\textbf{Parameter} & \textbf{Encoder} & \textbf{Longformer} \\
\hline
Optimizer & AdamW & AdamW \\
Learning rate & $5 \times 10^{-5}$ & $2 \times 10^{-5}$ \\
Weight decay & 0.01 & 0.01 \\
Batch size & 16 & 2 \\
Gradient accumulation & -- & 8 \\
Effective batch size & 16 & 16 \\
Max epochs & 20 & 20 \\
Early stopping patience & 5 & 7 \\
Max sequence length & 512 & 2048 \\
Attention window & -- & 512 \\
Dropout & 0.1$^*$ & 0.1 \\
Partial unfreezing & top $k$\% & top 25\% \\
\hline
\end{tabular}
\caption{Training configuration for encoder and Longformer models. 
$^*$Dropout $0.3$ is additionally evaluated in Stage~3.}
\label{tab:hyperparams}
\end{table}

\begin{table}[h]
\centering
\small
\begin{tabular}{ll}
\hline
\textbf{Dependency} & \textbf{Version} \\
\hline
Python & 3.12.7 \\
PyTorch & 2.5.1+cu118 \\
Transformers & 4.51.3 \\
NumPy & 2.3.5 \\
pandas & 2.2.3 \\
\hline
\end{tabular}
\caption{Software dependencies used in experiments.}
\label{tab:deps}
\end{table}

\section{Architecture and Model Variants}
\label{app:architecture}

\subsection{Classification Heads}
\label{app:heads}

We evaluate 3 classification head architectures:
\begin{itemize}
    \item \textbf{Default:} CLS $\rightarrow$ Dropout($0.1$) $\rightarrow$ Linear.
    \item \textbf{MLP:} CLS $\rightarrow$ Dropout($0.1$) $\rightarrow$ Linear $\rightarrow$ GELU $\rightarrow$ Dropout($0.1$) $\rightarrow$ Linear.
    \item \textbf{Average Pooling:} Mean pooling over non-padding token embeddings $\rightarrow$ Linear.
\end{itemize}

\subsection{Enriched Input Configuration}
\label{app:enriched}

In the enriched input setting, we concatenate the target question with context (i.e., the full interviewer turn) using explicit textual markers:

\begin{quote}
Target question: \textit{<question>} \\[4pt]
Full interviewer turn (context): \textit{<interview\_question>}
\end{quote}

This concatenated string forms the first sequence input to the encoder. 
The interview answer is provided as the second sequence.

\subsection{Encoder Final Configurations}
\label{app:encoder-configs}

The final configuration per encoder model after the four-stage optimization is as follows, ordered by dev macro-F1. All models selected weighted cross-entropy (WCE) loss and a dropout rate of $0.1$ as part of their optimal configuration.

\begin{enumerate}
    \item \texttt{DeBERTa-v3-base} (macro F1: $65.1$): Unfreeze 25\%, MLP head, head-tail truncation, augmented data, enriched input.
    \item \texttt{RoBERTa-base} (macro F1: $63.8$): Unfreeze 75\%, mean pooling head, head-tail truncation, original data, atomic input.
    \item \texttt{BERT-base} (macro F1: $63.4$): LoRA (rank~16), default head, head-tail truncation, original data, atomic input.
    \item \texttt{RoBERTa-large} (macro F1: $63.0$): Unfreeze 25\%, mean pooling head, standard truncation, original data, atomic input.
    \item \texttt{mBERT-base} (macro F1: $62.9$): Unfreeze 25\%, default head, standard truncation, augmented data, enriched input.
    \item \texttt{xlm-RoBERTa-base} (macro F1: $62.6$): Unfreeze 25\%, default head, standard truncation, augmented data, atomic input.
    \item \texttt{xlm-RoBERTa-large} (macro F1: $61.4$): Unfreeze 25\%, default head, head-tail truncation, original data, atomic input.
    \item \texttt{mDeBERTa-\allowbreak v3-\allowbreak base} (macro F1: $60.8$): Unfreeze 25\%, mean pooling head, standard truncation, original data, atomic input.
\end{enumerate}
\subsection{Longformer Configuration \& Ablation}
\label{app:longformer-configs}

\begin{figure}[h]
\centering
\includegraphics[width=\columnwidth]{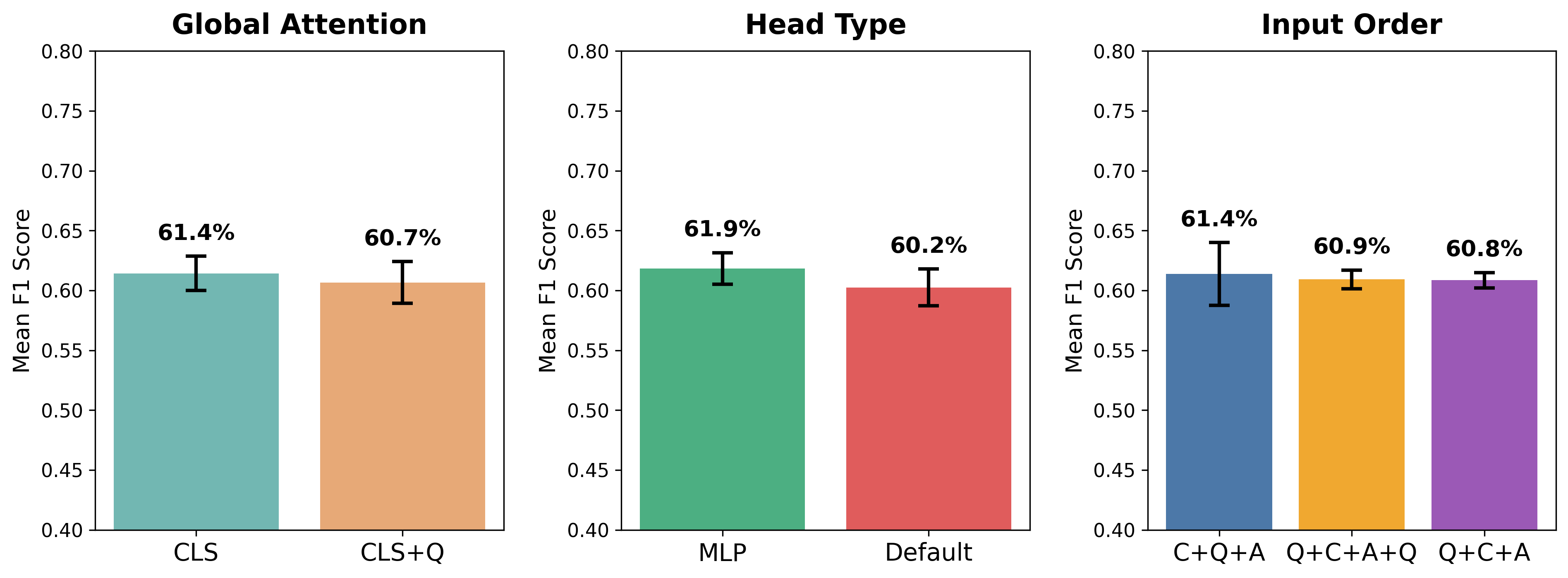}
\caption{Mean dev macro-F1 per design choice for the Longformer ablation. Error bars represent one standard deviation across configurations sharing each value.}
\label{fig:longformer-ablation}
\end{figure}

We evaluate 12 configurations across 3 design dimensions: input ordering, global attention pattern, and classification head type. All configurations use max sequence length 2048 and attention window 512. Figure~\ref{fig:longformer-ablation} reports mean dev macro-F1 per design choice, averaged over the remaining dimensions. Error bars represent one standard deviation across experiments sharing the same value of the given configuration dimension. The overlapping error bars across all 3 dimensions indicate that no single design choice yields a reliable improvement, and the results should be interpreted with caution. The results do not indicate a clear superior design choice, and further experimentation is needed.

\begin{table}[h]
\centering
\small
\begin{tabular}{lllc}
\hline
\textbf{Global Atten.} & \textbf{Head} & \textbf{Input Order} & \textbf{Dev F1} \\
\hline
\textbf{CLS}     & \textbf{MLP} & \textbf{C+Q+A}   & \textbf{64.3} \\
CLS+Q   & MLP & C+Q+A   & 62.6 \\
CLS+Q   & Default & Q+C+A+Q & 62.1 \\
CLS     & MLP & Q+C+A   & 61.9 \\
CLS     & Default  & C+Q+A   & 61.4 \\
CLS+Q   & MLP & Q+C+A+Q & 61.1 \\
CLS+Q   & MLP & Q+C+A   & 60.8 \\
CLS     & Default & Q+C+A   & 60.5 \\
CLS     & MLP & Q+C+A+Q & 60.4 \\
CLS+Q   & Default  & Q+C+A   & 60.2 \\
CLS     & Default  & Q+C+A+Q & 60.1 \\
CLS+Q   & Default  & C+Q+A   & 57.2 \\
\hline
\end{tabular}
\caption{Longformer ablation across 12 configurations, ordered by dev macro-F1. Input order reflects the token sequence fed to the model: C = full interviewer turn (context), Q = target question, A = answer. CLS+Q extends global attention to question tokens in addition to the CLS token.}
\label{tab:longformer-ablation}
\end{table}

\section{LLM Details}
\label{app:llm-configs}

\subsection{Model Checkpoints}
\label{app:llm-checkpoints} 
We report the exact Hugging Face checkpoints used for all LLM experiments to ensure reproducibility (Table~\ref{tab:llm-checkpoints}).

\begin{table}[h]
\centering
\small
\setlength{\tabcolsep}{4pt}
\begin{tabular}{ll}
\hline
\textbf{Model} & \textbf{Checkpoint} \\
\hline

\makecell[l]{LLaMA-3.1-Nemotron\\Ultra-253B} 
& \makecell[l]{Llama\_3\_1\_Nemotron\_Ultra\\253B\_v1} \\

LLaMA-3.3-70B-Instruct 
& \makecell[l]{Llama-3.3-70B\\Instruct} \\

Qwen3-235B-Instruct 
& \makecell[l]{Qwen3-235B-A22B\\Instruct-2507} \\

Qwen3-80B-Instruct 
& \makecell[l]{Qwen3-80B-A3B\\Instruct} \\

Qwen3-30B-Instruct 
& \makecell[l]{Qwen3-30B-A3B\\Instruct-2507} \\

Mixtral-8x22B-Instruct 
& \makecell[l]{Mixtral-8x22B\\Instruct} \\

\hline
\end{tabular}
\caption{Open-weight large language models (LLMs) used in our experiments, with their corresponding checkpoints.}
\label{tab:llm-checkpoints}
\end{table}

\subsection{Prompt Templates}
\label{app:prompts}

All Stage~1 prompting strategies (see Section~\ref{sec:task1-prompting}) share a common system role and label definitions, shown in the Zero-Shot (ZS) template below. 
Subsequent strategies extend this base prompt (ZS+Re2 repeats the instruction block; FS prepends demonstrations; CoT-inspired prompts add internal reasoning instructions). 
Stage~2 prompt optimization refines the best-performing Stage~1 configuration (FS27) by expanding Clarity label definitions with Level-2 (i.e., Evasion) subcategory descriptions and, in some variants, prepending the president’s name.
Input placeholders \texttt{\{Q\}} and \texttt{\{A\}} are replaced at inference time with the target question and answer, respectively. In the enriched input setting, \texttt{\{Q\}} is replaced with the concatenated target question and full interviewer turn as described in Appendix~\ref{app:enriched}.
\paragraph{Zero-Shot (ZS).} Defines the role, label set, and output format. All subsequent strategies extend this base prompt.

% \begin{quote}
% \small
% \ttfamily
% You are a world-class political discourse analyst trained to detect evasive 
% communication strategies in high-stakes interviews.

% Judge the clarity of an answer to a journalist's question.

% Labels (choose exactly ONE):\\
% - Clear Reply: the answer directly supplies what was asked.\\
% - Ambivalent Reply: the answer references the topic but is indirect, vague, 
% partial, or hedged.\\
% - Clear Non-Reply: the answer refuses, claims not to know, asks for 
% clarification, or ignores the question.

% Output ONLY the label name (no explanation, no punctuation).

% QUESTION: \{Q\}\\
% ANSWER: \{A\}\\
% Label:
% \end{quote}

%%%%%%%%%%%%%%%%%%%%%%%%%%%
\begin{tcolorbox}[breakable, colback=gray!5, colframe=black!40, boxrule=0.5pt, arc=2pt, left=5pt, right=5pt, top=5pt, bottom=5pt, width=0.95\linewidth]

\ttfamily

You are a world-class political discourse analyst trained to detect evasive communication strategies in high-stakes interviews.\\

Judge the clarity of an answer to a journalist's question.\\

Labels (choose exactly ONE):\\
- Clear Reply: the answer directly supplies what was asked.\\
- Ambivalent Reply: the answer references the topic but is indirect, vague, partial, or hedged.\\
- Clear Non-Reply: the answer refuses, claims not to know, asks for 
clarification, or ignores the question.\\

Output ONLY the label name (no explanation, no punctuation).\\

QUESTION: \{Q\}\\
ANSWER: \{A\}\\
Label:

\end{tcolorbox}
%%%%%%%%%%%%%%%%%%%%%%%%%%%

\paragraph{Zero-Shot with Instruction Repetition (ZS+Re2).}
Extends ZS by repeating the instruction block before the input.

% \begin{quote}
% \small\ttfamily
% You are a world-class political discourse analyst trained to detect evasive 
% communication strategies in high-stakes interviews.

% Judge the clarity of an answer to a journalist's question.

% Labels (choose exactly ONE):\\
% - Clear Reply: the answer directly supplies what was asked.\\
% - Ambivalent Reply: the answer references the topic but is indirect, vague, 
% partial, or hedged.\\
% - Clear Non-Reply: the answer refuses, claims not to know, asks for 
% clarification, or ignores the question.

% Output ONLY the label name (no explanation, no punctuation).

% Read the question again:\\
% Judge the clarity of an answer to a journalist's question.

% Labels (choose exactly ONE):\\
% - Clear Reply: the answer directly supplies what was asked.\\
% - Ambivalent Reply: the answer references the topic but is indirect, vague, 
% partial, or hedged.\\
% - Clear Non-Reply: the answer refuses, claims not to know, asks for 
% clarification, or ignores the question.

% Output ONLY the label name (no explanation, no punctuation).

% QUESTION: \{Q\}\\
% ANSWER: \{A\}\\
% Label:
% \end{quote}

%%%%%%%%%%%%%%%%%%%%%%%%%%%
\begin{tcolorbox}[breakable, colback=gray!5, colframe=black!40, boxrule=0.5pt, arc=2pt, left=5pt, right=5pt, top=5pt, bottom=5pt, width=0.95\linewidth]

\ttfamily

You are a world-class political discourse analyst trained to detect evasive communication strategies in high-stakes interviews.\\

Judge the clarity of an answer to a journalist's question.\\

Labels (choose exactly ONE):\\
- Clear Reply: the answer directly supplies what was asked.\\
- Ambivalent Reply: the answer references the topic but is indirect, vague, 
partial, or hedged.\\
- Clear Non-Reply: the answer refuses, claims not to know, asks for 
clarification, or ignores the question.\\

Output ONLY the label name (no explanation, no punctuation).\\

Read the question again:\\
Judge the clarity of an answer to a journalist's question.

Labels (choose exactly ONE):\\
- Clear Reply: the answer directly supplies what was asked.\\
- Ambivalent Reply: the answer references the topic but is indirect, vague, 
partial, or hedged.\\
- Clear Non-Reply: the answer refuses, claims not to know, asks for 
clarification, or ignores the question.\\

Output ONLY the label name (no explanation, no punctuation).\\

QUESTION: \{Q\}\\
ANSWER: \{A\}\\
Label:

\end{tcolorbox}
%%%%%%%%%%%%%%%%%%%%%%%%%%%

\paragraph{Few-Shot (FS).}
Few-shot prompts extend the ZS base by inserting $k$ labelled demonstrations
between the label definitions and the input. Demonstrations are class-balanced
(equal number per class) and drawn from the training set. We evaluate $k \in
\{3, 9, 27\}$ shots.
The structure is as follows:

% \begin{quote}
% \small\ttfamily
% You are a world-class political discourse analyst trained to detect evasive
% communication strategies in high-stakes interviews.

% Judge the clarity of an answer to a journalist's question.

% Labels (choose exactly ONE):\\
% - Clear Reply: the answer directly supplies what was asked.\\
% - Ambivalent Reply: the answer references the topic but is indirect, vague,
% partial, or hedged.\\
% - Clear Non-Reply: the answer refuses, claims not to know, asks for
% clarification, or ignores the question.

% The following examples illustrate each label:

% Example 1:\\
% Question: \{example\_question\}\\
% Answer: \{example\_answer\}\\
% Label: \{label\}
% \\

% \textrm{[$k$ class-balanced demonstrations]}\\

% Output ONLY the label name (no explanation, no punctuation).

% Question: \{Q\}\\
% Answer: \{A\}\\
% Label:
% \end{quote}

%%%%%%%%%%%%%%%%%%%%%%%%%%%
\begin{tcolorbox}[breakable, colback=gray!5, colframe=black!40, boxrule=0.5pt, arc=2pt, left=5pt, right=5pt, top=5pt, bottom=5pt, width=0.95\linewidth]
\ttfamily

You are a world-class political discourse analyst trained to detect evasive communication strategies in high-stakes interviews.\\

Judge the clarity of an answer to a journalist's question.\\

Labels (choose exactly ONE):\\
- Clear Reply: the answer directly supplies what was asked.\\
- Ambivalent Reply: the answer references the topic but is indirect, vague,
partial, or hedged.\\
- Clear Non-Reply: the answer refuses, claims not to know, asks for
clarification, or ignores the question.\\

The following examples illustrate each label:\\

Example 1:\\
Question: \{example\_question\}\\
Answer: \{example\_answer\}\\
Label: \{label\}
\\

\textrm{[$k$ class-balanced demonstrations]}\\

Output ONLY the label name (no explanation, no punctuation).\\

Question: \{Q\}\\
Answer: \{A\}\\
Label:

\end{tcolorbox}
%%%%%%%%%%%%%%%%%%%%%%%%%%%

\paragraph{CoT-Inspired (CoT).}
The CoT-inspired prompt replaces the output instruction of the ZS base with
an internal reasoning directive. Unlike standard CoT, the model is instructed
to reason internally without outputting any intermediate steps:

% \begin{quote}
% \small\ttfamily
% You are a world-class political discourse analyst trained to detect evasive
% communication strategies in high-stakes interviews.

% Judge the clarity of an answer to a journalist's question.

% Labels (choose exactly ONE):\\
% - Clear Reply: the answer directly supplies what was asked.\\
% - Ambivalent Reply: the answer references the topic but is indirect, vague,
% partial, or hedged.\\
% - Clear Non-Reply: the answer refuses, claims not to know, asks for
% clarification, or ignores the question.

% In your own mind, reason step by step about how the answer responds to
% the question. Keep all intermediate reasoning hidden and do not write it out.\\
% When you have decided, output ONLY the label name (no explanation, no
% extra text, no punctuation).

% QUESTION: \{Q\}\\
% ANSWER: \{A\}\\
% Label:
% \end{quote}

%%%%%%%%%%%%%%%%%%%%%%%%%%%
\begin{tcolorbox}[breakable, colback=gray!5, colframe=black!40, boxrule=0.5pt, arc=2pt, left=5pt, right=5pt, top=5pt, bottom=5pt, width=0.95\linewidth]
\ttfamily

You are a world-class political discourse analyst trained to detect evasive
communication strategies in high-stakes interviews.\\

Judge the clarity of an answer to a journalist's question.\\

Labels (choose exactly ONE):\\
- Clear Reply: the answer directly supplies what was asked.\\
- Ambivalent Reply: the answer references the topic but is indirect, vague,
partial, or hedged.\\
- Clear Non-Reply: the answer refuses, claims not to know, asks for
clarification, or ignores the question.\\

In your own mind, reason step by step about how the answer responds to
the question. Keep all intermediate reasoning hidden and do not write it out.\\

When you have decided, output ONLY the label name (no explanation, no
extra text, no punctuation).\\

QUESTION: \{Q\}\\
ANSWER: \{A\}\\
Label:
\end{tcolorbox}
%%%%%%%%%%%%%%%%%%%%%%%%%%%

\paragraph{Stage~2: Subcategory-Augmented Optimized Prompt.}
Building on FS27, Stage~2 replaces the Clarity label definitions with expanded
versions that include Level-2 Evasion subcategories. The user message includes the target question, the context (full interviewer turn), and the answer, and optionally prepends the speaker's name.

\begin{tcolorbox}[breakable, colback=gray!5, colframe=black!40, boxrule=0.5pt, arc=2pt, left=5pt, right=5pt, top=5pt, bottom=5pt, width=0.95\linewidth]
\ttfamily

You are a world-class political discourse analyst trained to detect evasive communication strategies in high-stakes interviews.\\

Judge the clarity of an answer to a journalist's question.\\

Labels (choose exactly ONE):\\
Ambivalent Reply\\
Definition: Where a response is given in the form of a valid answer but
allows for multiple interpretations.\\
Sub-categories:\\
- Implicit: The information requested is given, but without being
explicitly stated (not in the expected form)\\
- General: The information provided is too general/lacks the requested
specificity\\
- Partial: Offers only a specific component of the requested information\\
- Dodging: Ignoring the question altogether\\
- Deflection: Starts on topic but shifts the focus and makes a different
point than what is asked\\[4pt]
\\
Clear Reply\\
Definition: Containing replies that admit only one interpretation.\\
Sub-category:\\
- Explicit: The information requested is explicitly stated (in the
requested form)\\[4pt]
\\
Clear Non-Reply\\
Definition: Containing responses where the answerer openly refuses to
share information.\\
Sub-categories:\\
- Declining to answer: Acknowledge the question but directly or
indirectly refusing to answer at the moment\\
- Claims ignorance: The answerer claims/admits not to know the answer
themselves\\
- Clarification: Does not provide the requested information and asks
for clarification\\

The following examples illustrate each category:\\

\textrm{[27 class-balanced demonstrations]}\\

Output ONLY the category name.\\

Target question (to evaluate): \{Q\}\\
Speaker: \{president\}\\
Full interviewer turn (may contain multiple questions): \{context\}\\
Answer: \{A\}\\
Label:
\end{tcolorbox}
%%%%%%%%%%%%%%%%%%%%%%%%%%%

% The 27 demonstrations and output instruction remain identical to FS27 as shown below (note that when president conditioning is disabled, the \texttt{Speaker} line is omitted).

% %%%%%%%%%%%%%%%%%%%%%%%%%%%
% \begin{tcolorbox}[breakable, colback=gray!5, colframe=black!40, boxrule=0.5pt, arc=2pt, left=5pt, right=5pt, top=5pt, bottom=5pt, width=0.95\linewidth]

% Target question (to evaluate): \{Q\}\\
% Speaker: \{president\}\\
% Full interviewer turn (may contain multiple questions): \{context\}\\
% Answer: \{A\}\\
% Label:
% \end{tcolorbox}
% %%%%%%%%%%%%%%%%%%%%%%%%%%%

\paragraph{Task~2: Few-Shot Evasion Classification (FS27)}

Task~2 mirrors the optimized Task~1 Stage~2 prompt (FS27, enriched input, subcategory-augmented definitions), adapted to the 9 Evasion categories. Demonstrations are class-balanced across the 9 Evasion categories and drawn from the training split. The structure is as follows:

% \begin{quote}
% \small
% \ttfamily
% You are a world-class political discourse analyst trained to detect evasive
% communication strategies in high-stakes interviews.

% Your task is to classify how a public official answers a journalist’s question.

% Labels (choose exactly ONE):\\
% - Implicit: The information requested is given, but without being explicitly stated (not in the expected form).\\
% - General: The information provided is too general or lacks the requested specificity.\\
% - Partial/half-answer: Offers only a specific component of the requested information.\\
% - Dodging: Ignoring the question altogether.\\
% - Deflection: Starts on topic but shifts the focus and makes a different point than what is asked.\\
% - Explicit: The information requested is explicitly stated (in the requested form).\\
% - Declining to answer: Acknowledges the question but directly or indirectly refuses to answer.\\
% - Claims ignorance: The answerer claims or admits not to know the answer.\\
% - Clarification: Does not provide the requested information and asks for clarification.

% The following examples illustrate each category:

% Example 1:\\
% Question: \{example\_question\}\\
% Answer: \{example\_answer\}\\
% Label: \{label\}\\

% \textrm{[27 class-balanced demonstrations]}\\

% Output ONLY the category name.

% Question: \{Q\}\\
% Answer: \{A\}\\
% Label:
% \end{quote}

%%%%%%%%%%%%%%%%%%%%%%%%%%%
\begin{tcolorbox}[breakable, colback=gray!5, colframe=black!40, boxrule=0.5pt, arc=2pt, left=5pt, right=5pt, top=5pt, bottom=5pt, width=0.95\linewidth]
\ttfamily

You are a world-class political discourse analyst trained to detect evasive communication strategies in high-stakes interviews.\\

Your task is to classify how a public official answers a journalist’s question.\\

Labels (choose exactly ONE):\\
- Implicit: The information requested is given, but without being explicitly stated (not in the expected form).\\
- General: The information provided is too general or lacks the requested specificity.\\
- Partial/half-answer: Offers only a specific component of the requested information.\\
- Dodging: Ignoring the question altogether.\\
- Deflection: Starts on topic but shifts the focus and makes a different point than what is asked.\\
- Explicit: The information requested is explicitly stated (in the requested form).\\
- Declining to answer: Acknowledges the question but directly or indirectly refuses to answer.\\
- Claims ignorance: The answerer claims or admits not to know the answer.\\
- Clarification: Does not provide the requested information and asks for clarification.\\

The following examples illustrate each category:\\

% Example 1:\\
% Question: \{example\_question\}\\
% Answer: \{example\_answer\}\\
% Label: \{label\}\\

\textrm{[27 class-balanced demonstrations]}\\

Output ONLY the category name.\\

Target question (to evaluate): \{Q\}\\
Speaker: \{president\}\\
Full interviewer turn (may contain multiple questions): \{context\}\\
Answer: \{A\}\\
Label:

\end{tcolorbox}
%%%%%%%%%%%%%%%%%%%%%%%%%%%

\subsection{Stage 1: Technique \& Model Sweep: Best Configurations per LLM}
\label{app:llm-stage1}

Table~\ref{tab:llm-stage1} reports the best Stage~1 configuration per LLM on the dev set, ordered by dev macro-F1.

\begin{table}[h]
\centering
\small
\begin{tabular}{lcc}
\hline
\textbf{Model} & \textbf{Strategy} & \textbf{Dev F1} \\
\hline
\texttt{Qwen3-\allowbreak 235B-\allowbreak Instruct} & FS27 & 67.2 \\
\texttt{Qwen3-\allowbreak 80B-\allowbreak Instruct} & FS3 & 64.8 \\
\texttt{GPT-\allowbreak 5} & FS27 & 64.4 \\
\texttt{LLaMA-\allowbreak 3.3-\allowbreak 70B-\allowbreak Instruct} & FS27 & 62.5 \\
\texttt{Mixtral-\allowbreak 8x22B-\allowbreak Instruct} & ZS+Re2 & 58.5 \\
\texttt{Qwen3-\allowbreak 30B-\allowbreak Instruct} & FS27 & 58.1 \\
\texttt{LLaMA-\allowbreak 3.1-\allowbreak Nemotron-\allowbreak Ultra-\allowbreak 253B} & FS27 & 43.5 \\
\hline
\end{tabular}
\caption{Best Stage~1 configuration per LLM. All models use enriched input.}
\label{tab:llm-stage1}
\end{table}

\subsection{Stage 2: Prompt Optimization \& Model Expansion: Model Configurations}

The two top Stage~1 models were both Qwen variants (see Table~\ref{tab:llm-stage1}, Stage~1 results). To ensure model family diversity, we retain \texttt{Qwen3-\allowbreak 235B} ($67.2$) and \texttt{GPT-\allowbreak 5} ($64.4$) and expand to \texttt{Gemini-\allowbreak 3-\allowbreak Flash-\allowbreak Preview} and \texttt{Claude-\allowbreak Opus-\allowbreak 4.5}. All models use FS27 with enriched input and subcategory-augmented label definitions. The top 3 after Stage~2 (Table~\ref{tab:llm-stage2}) were chosen to form the LLM ensemble; \texttt{Claude-\allowbreak Opus-\allowbreak 4.5} was not among the top 3 and was not retained.

\begin{table}[h]
\centering
\small
\begin{tabular}{lcc}
\hline
\textbf{Model} & \textbf{President} & \textbf{Dev F1} \\
\hline
\texttt{Gemini-3-Flash-Preview}  & Yes & 71.9 \\
\texttt{GPT-5}                  & No  & 71.5 \\
\texttt{Qwen3-235B-Instruct}     & Yes & 68.3 \\
\hline
\end{tabular}
\caption{Top 3 Stage~2 results per LLM (FS27, enriched input, subcategory-augmented 
label definitions). \textbf{President} indicates whether the president's name 
was included in the prompt. These 3 models form the final ensemble.}
\label{tab:llm-stage2}
\end{table}

\section{Data Augmentation}
\label{app:augmentation}

\subsection{Augmentation Procedure}
\label{app:aug-procedure}
To mitigate class imbalance, as part of Stage~4, we augment the minority class in Task~1 (\textit{Clear Non-Reply}) by generating paraphrases using \texttt{GPT-4o} ($\text{temperature} = 0.55$). Starting from 356 \textit{Clear Non-Reply} instances in the training set, we generate one paraphrase per instance (356 total), followed by an additional 194 paraphrases sampled uniformly at random from the same instances, yielding 550 new candidates in total. This oversampling accounts for expected filtering losses. After post-generation filtering (see Appendix~\ref{app:aug-filtering}), 356 valid paraphrases (from the 550 new instances) are retained and appended to the training set, increasing the minority class from 356 to 712 instances.
% To mitigate class imbalance, we augment the minority (\textit{Clear Non-Reply}) class in Task 1 by generating paraphrases with qwen-4o (temperature = 0.55). From the 356 training instances, we generate one paraphrase per instance plus 194 sampled uniformly at random, yielding 550 new candidates in total. This oversampling accounts for expected losses from the post-generation filtering described below.

\subsection{Prompt-Level Constraints}
\label{app:aug-constraints}
The paraphrase generation prompt imposes a set of structural and behavioral constraints. Each rewritten answer must remain a \textit{Clear Non-Reply}: it must not directly address the interviewer’s question and must retain its deflective, no-answering intent. The prompt further requires preserving the original tone, level of formality, and stylistic markers (e.g., vagueness, hesitation); avoiding the introduction of new facts, examples, or topics; and maintaining the original discourse structure without summarizing, clarifying, or restructuring the response.

\subsection{Post-Generation Filtering}
\label{app:aug-filtering}
Each generated paraphrase is kept only if it satisfies both conditions relative to its source: (1) the total word count within $\pm 20\%$ of the original, and (2) the sentence count within $\pm 1$ of the original. Generation of paraphrases is retried up to 3 times per instance. Only paraphrases passing both filters are eligible for inclusion in the final training set. From the filtered pool, the first 356 valid paraphrases are appended to the original training data, doubling the size of the minority class.

\section{Confusion Matrices}
\label{app:confusion-matrices}

\begin{figure}[h]
\centering
\includegraphics[width=\columnwidth]{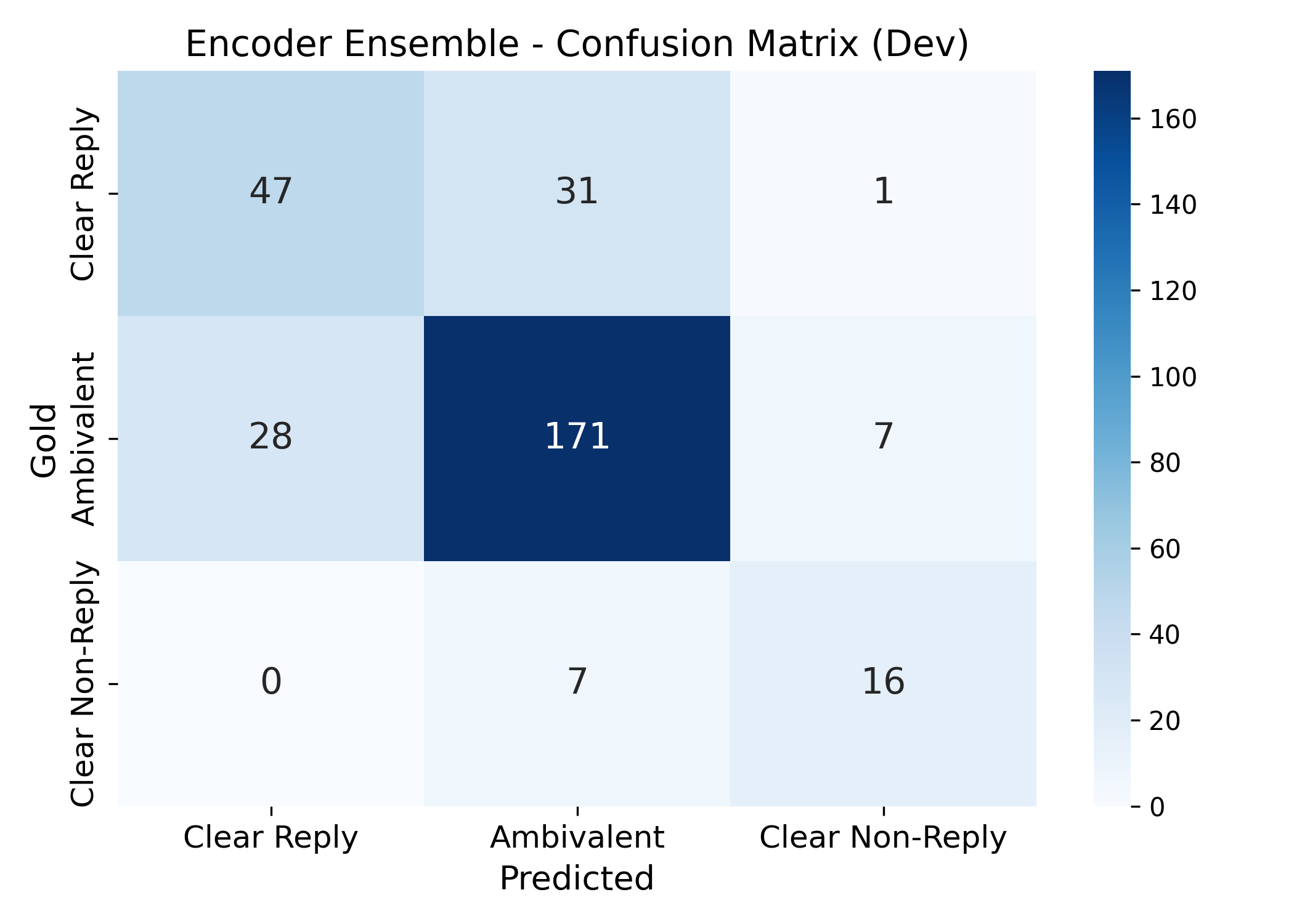}
\caption{Encoder ensemble confusion matrix (dev set).}
\label{fig:confusion-matrix-encoder}
\end{figure}

\begin{figure}[h]
\centering
\includegraphics[width=\columnwidth]{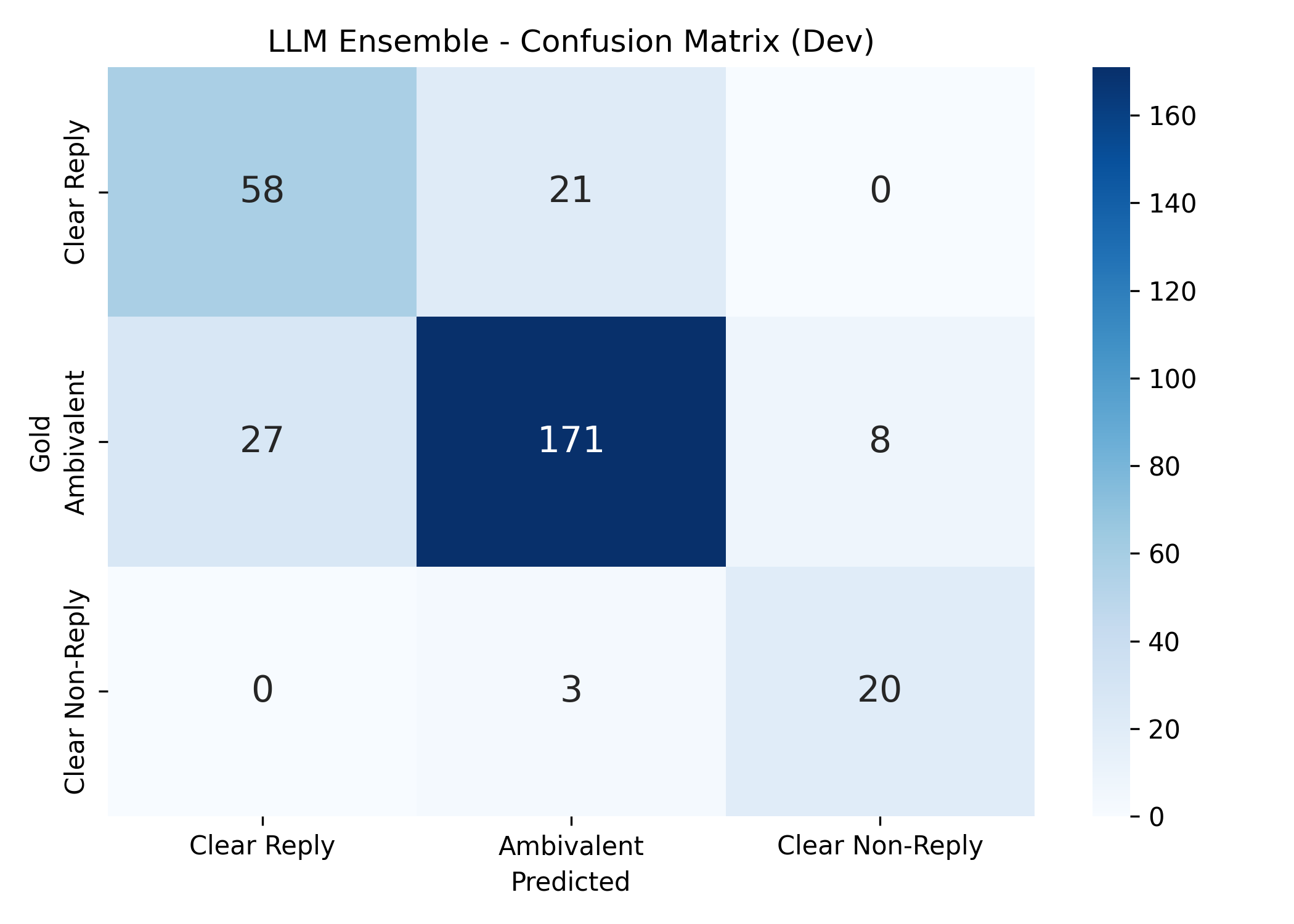}
\caption{LLM ensemble confusion matrix (dev set).}
\label{fig:confusion-matrix-llm}
\end{figure}

% \subsection{Encoder Ensemble}
% \label{app:confusion-matrix-encoder}
Both the encoder and LLM ensembles (Figures~\ref{fig:confusion-matrix-encoder} and~\ref{fig:confusion-matrix-llm}) share the same dominant error pattern: bidirectional confusion between \textit{Clear Reply} and \textit{Ambivalent} ($28 + 31 = 59$ instances for the encoder, $21 + 27 = 48$ for the LLM ensemble). This mirrors the lowest pairwise annotator agreement among Clarity classes ($\kappa = 0.65$; \citealt{thomas-etal-2024-never}), suggesting this boundary is inherently difficult for both humans and models. The LLM ensemble improves \textit{Clear Non-Reply} recall ($20/23$ vs.\ $16/23$), consistent with its stronger minority class performance reported in Table~\ref{tab:llm_vs_encoder}.

% \subsection{LLM Ensemble}
% \label{app:confusion-matrix-llm}

\end{document}